\documentclass[10pt, a4paper]{article}

\usepackage{tabularx}
\newcolumntype{Y}{>{\centering\arraybackslash}X}
\usepackage{float}
\usepackage{afterpage}
\usepackage{siunitx}

\usepackage{lrec-coling2024} 

\usepackage{enumitem}
\title{Could We Have Had Better Multilingual LLMs If English Was Not the Central Language?\\ \vspace*{.5\baselineskip}}

\name{Ryandito Diandaru$^{\spadesuit}$, Lucky Susanto$^{\diamondsuit}$, Zilu Tang$^{\clubsuit}$, \\ {\bf \large Ayu Purwarianti$^{\spadesuit}$, Derry Wijaya$^{\clubsuit,\heartsuit}$}}

\address{Bandung Institute of Technology$^{\spadesuit}$, University of Indonesia$^{\diamondsuit}$, Boston University$^{\clubsuit}$,\\Monash University Indonesia$^{\heartsuit}$\\
         13519157@std.stei.itb.ac.id, lucky.susanto@ui.ac.id, zilutang@bu.edu, ayu@itb.ac.id,\\ derry.wijaya@monash.edu}

\abstract{
Large Language Models (LLMs) demonstrate strong machine translation capabilities on languages they are trained on. However, the impact of factors beyond training data size on translation performance remains a topic of debate, especially concerning languages not directly encountered during training. Our study delves into Llama2's translation capabilities. By modeling a linear relationship between linguistic feature distances and machine translation scores, we ask ourselves if there are potentially better central languages for LLMs other than English. Our experiments show that the 7B Llama2 model yields above 10  BLEU when translating into all languages it has seen, which rarely happens for languages it has not seen. Most translation improvements into unseen languages come from scaling up the model size rather than instruction tuning or increasing shot count. Furthermore, our correlation analysis reveals that syntactic similarity is not the only linguistic factor that strongly correlates with machine translation scores. Interestingly, we discovered that under specific circumstances, some languages (e.g. Swedish, Catalan), despite having significantly less training data, exhibit comparable correlation levels to English. These insights challenge the prevailing landscape of LLMs, suggesting that models centered around languages other than English could provide a more efficient foundation for multilingual applications.
 \\ \newline  }

\begin{document}

\maketitleabstract

\section{Introduction}

Large Language Models (LLMs) have been a popular research topic in Natural Language Processing (NLP) due to their remarkable performance on various tasks including machine translation \cite{brown2020language, openai2023gpt4, touvron2023llama, touvron2023llama2}. Extensive evaluations on machine translation of the popular GPT model family \cite{openai2023gpt4} have suggested that they can translate high-resource languages \cite{robinson2023chatgpt, hendy2023good}. However, it is rarely the case for low-resource or underrepresented languages \cite{robinson2023chatgpt, hendy2023good, stap-araabi-2023-chatgpt, kadaoui2023tarjamat}.

A straightforward approach for the lack of training data in low-resource translation is to collect more labeled data. However, investing in data creation is nontrivial as it comes with challenges, including the cost of such endeavors. For example, \citet{aji-etal-2022-one} described the absence of Wikipedia articles on Indonesian regional languages and the challenges of labeled data collection for them, which includes the lack of speakers, the diversity of dialects, and the lack of a writing standard. In addition, training large language models on more data brings environmental consequences \cite{strubell-etal-2019-energy}. In the long run, more training data may require longer GPU compute hours, which will release more greenhouse gas emissions. 

Aside from data creation, other techniques are often employed as an alternative. A popular approach for multilingual or low-resource NLP is to leverage other languages to benefit from cross-lingual transfer. These approaches include using them as pivot \cite{wijaya-etal-2017-learning,xia-etal-2019-generalized}, transfer learning \cite{gu-etal-2018-universal, nguyen-chiang-2017-transfer}, and joint training \cite{neubig-hu-2018-rapid, johnson-etal-2017-googles}. Improvements from such methods indicate a strong influence of the presence of other languages in the training data. Given that including related languages alongside the low-resource language can improve performance \cite{xia-etal-2019-generalized, poncelas-effendi-2022-benefiting, gu-etal-2018-universal, nguyen-chiang-2017-transfer, neubig-hu-2018-rapid, johnson-etal-2017-googles}, it is beneficial to include proximity measurements between these languages on evaluations, which can be done using the vectors from the URIEL database \cite{littell-etal-2017-uriel}. The utilization of the URIEL database has made evaluating multiple languages more explainable by leveraging linguistically aware feature vectors from which linguistic distances can be computed. These vectors have been utilized by previous works in various ways including determining which language to use as transfer or pivot language \cite{lin-etal-2019-choosing, nambi2023breaking} and measuring language diversity \cite{ruder-etal-2021-xtreme}.

It has been established that there are benefits to using other languages in the training process. However, multilingual labeled data creation is challenging. In this paper, we aim to provide hints to narrow down future data collection strategies by evaluating an existing LLM family. A constraint in previous studies that assess the GPT model series \cite{hendy2023good, robinson2023chatgpt} has been the fact that these models are proprietary, closed-source systems that do not disclose information regarding their training data. This presents a challenge as it remains unclear which languages are included in the training of the models. On the other hand, open-source LLMs such as Meta's Llama2 \cite{touvron2023llama2}, is more transparent about its training process, including the languages that are included in its training data. This makes the model more suitable as a subject for our evaluation.

In this work, we are evaluating Llama2 \cite{touvron2023llama2} for machine translation to highlight its multilingual capability in languages it has or has not seen during training. We also model a linear relationship (through correlation scores) between the linguistic feature distances and the translation metrics and use these scores as a basis for language importance analysis. The goal of the analysis is to narrow down the data investment effort by shedding light on which language(s) may improve the translation of other languages when included in the training data. 
An efficient data collection strategy will result in future multilingual LLMs that can be trained and deployed more efficiently, thus promoting sustainability. In summary, our contributions are as follows:
\begin{enumerate}[noitemsep,topsep=0pt]
\item We evaluate Llama2 and provide machine translation scores of this model for 41 languages, 15 of which were not seen during its training.
\item We reveal that increasing model parameters is more effective in improving translation over instruction tuning and few-shot learning.
\item Our research reveals that syntactic similarity between languages is not the only linguistic aspect that is strongly linked to machine translation performance. Surprisingly, these strong correlations between linguistic feature distances and machine translation performances extend beyond English and hold true across various languages, therefore opening up the possibility of other better central languages for multilingual LMs
\end{enumerate}

\section{Methodology}
\subsection{Machine Translation Evaluation}
We experiment with languages reported in the training data of Llama2 \cite{touvron2023llama2}, the list of which and their respective ISO 639-3 codes can be found in Table \ref{tab:inllama-languages}. We refer to this set of languages as \textbf{inllama}. We also pick 15 languages not reported in the training data which we will refer to as \textbf{outllama}, presented in Table \ref{tab:outllama-languages}. It is important to highlight that languages not explicitly mentioned in Llama2 might still be present in the training data, albeit at a minuscule proportion of less than 0.005\% of its training data \cite{touvron2023llama2}. Languages in \textbf{outllama} cover various language genera and writing systems. The machine translation evaluation is conducted using the FLORES-200 \citelanguageresource{guzman-etal-2019-flores} benchmark as it is available for numerous low-resource languages. We exclude X$\rightarrow$English translation directions to mitigate the risk of potential data leakage, given that FLORES-200 uses Wikipedia for its English sentences. We also exclude zero-shot translation as LLMs often get the language wrong in this prompting setup as reported by \citet{robinson2023chatgpt}. We measure translation quality using machine translation scores. Translation quality is measured with the BLEU score \cite{papineni-etal-2002-bleu} and a model-based machine translation metric (COMET-22 \cite{rei-etal-2022-comet}) where applicable. 
COMET-22 is used to compensate for the drawbacks of BLEU and vice-versa. 

\begin{table}
    \small
    \centering
    \begin{tabularx}{\columnwidth}{lYYY} 
    \hline
         \textbf{Language}& \textbf{Genus}& \textbf{BLEU}&\textbf{COMET-22}\\ 
    \hline
         German (deu)& Germanic & 33.68 & 0.83\\
 Swedish (swe)& Germanic &37.71 &0.87\\
 Dutch (nld)& Germanic & 27.45&0.84\\
 Norwegian (nor)& Germanic & 29.54&0.86\\
 Danish (dan)& Germanic & 36.21&0.86\\
 French (fra)& Romance & 42.4&0.84\\
 Spanish (spa)& Romance & 28.54&0.84\\
 Italian (ita)& Romance & 28.78&0.85\\
 Portuguese (por)& Romance & 43.21&0.87\\
 Catalan (cat)& Romance & 35.92&0.84\\
 Romanian (ron)& Romance & 31.58&0.84\\
 Russian (rus)& Slavic & 28.21&0.85\\
 Polish (pol)& Slavic & 22.34&0.83\\
 Ukrainian (ukr)& Slavic & 26.03&0.83\\
 Serbian (srp)& Slavic & 23.96&0.81\\
 Czech (ces)& Slavic & 24.94&0.82\\
 Bulgarian (bul)& Slavic & 29.57&0.83\\
 Croatian (hrv)& Slavic & 21.3&0.81\\
 Slovenian (slv)& Slavic & 19.51&0.77\\
 Chinese (zho)& Chinese & 19.79&0.82\\
 Japanese (jpn)& Japanese & 17.02&0.84\\
 Vietnamese (vie)& Vietic & 28.77&0.82\\
 Korean (kor)& Korean & 11.08&0.78\\
 Indonesian (ind)& Malayo-Sumbawan & 31.15&0.86\\
 Finnish (fin)& Finnic & 18.08&0.82\\
 Hungarian (hun)& Ugric & 18.4&0.78\\
 \hline
    \end{tabularx}
    \caption{List of \textbf{inllama} languages along with their ISO 639-3 codes, genus, and machine translation scores obtained using one-shot Llama2-7B.}
    \label{tab:inllama-languages}
\vspace{-5mm}
\end{table}

We aim to experiment with open-source LLMs that replicate proprietary models such as ChatGPT  \cite{openai2023gpt4} in terms of usability and safety. At the time the Llama2 model was released and the experiment design for this paper was constructed, none of the open-source models are suitable substitutes for production models as they may not have been aligned to match human preferences and there may be a performance gap \cite{touvron2023llama2}. On account of this, we decided to move forward only with the Llama2 model family.
The machine translation evaluation begins with one-shot translations for both languages in \textbf{inllama} and \textbf{outllama} using the vanilla 7B model. From this experiment, we categorize languages that yield under 10 BLEU as \textbf{unlearned} languages\footnote{Based on "Almost useless" interpretation from \url{https://cloud.google.com/translate/automl/docs/evaluate}}.
For the \textbf{unlearned} languages, we experiment further with model scale, chat version, and adding the shot count to maximize the potential of in-context learning. Our choice of randomly picking 5 shots from the validation set of FLORES-200 is motivated by the experimental setup used by \citet{hendy2023good} which states that increasing beyond 5 shots does not result in meaningful improvement and shows that selected quality shots do not always improve more than 1 BLEU compared to random selections for GPT (text-davinci-003) model. For translation with chat models, we use the prompt by \citet{robinson2023chatgpt} which follows the recommendation of \citet{gao2023design} for designing prompts for translation using instruction-tuned models. The prompts used in our experiments are given in Table \ref{tab:prompts}
\begin{table}
    \small
    \centering
    \begin{tabularx}{\columnwidth}{lYY}
    \hline
         \textbf{Language}& \textbf{Genus} & \textbf{Writing System}\\
    \hline
         Afrikaans (afr)& Germanic & Latin\\
         Galician (glg)& Romance & Latin\\
         Macedonian (mkd)& Slavic & Cyrillic\\
         Slovak (slk)& Slavic & Latin\\
         Armenian (hye)& Armenian & Armenian\\
         Basque (eus)& Basque & Latin\\
         Georgian (kat)& Kartvelian & Georgian\\
         Icelandic (isl)& Germanic & Latin\\
         Igbo (ibo)& Igboid & Latin\\
 Javanese (jav)& Javanese & Latin\\
 Sinhala (sin)& Indic & Sinhala\\
 Tagalog (tgl)& Greater Central Philippine & Latin\\
 Tamil (tam)& Dravidian & Tamil\\
 Telugu (tel)& Dravidian & Telugu\\
 Welsh (cym)& Celtic & Latin\\
 \hline
    \end{tabularx}
    \caption{List of \textbf{outllama} languages and their ISO 639-3 codes. We also include in this table additional language information retrieved from WALS \cite{wals} }
    \label{tab:outllama-languages}
\vspace{-5mm}
\end{table}

\subsection{Correlation Score Analysis}\label{corr-sc-analysis}
\begin{figure*}
    \centering
    \includegraphics[width=\textwidth]{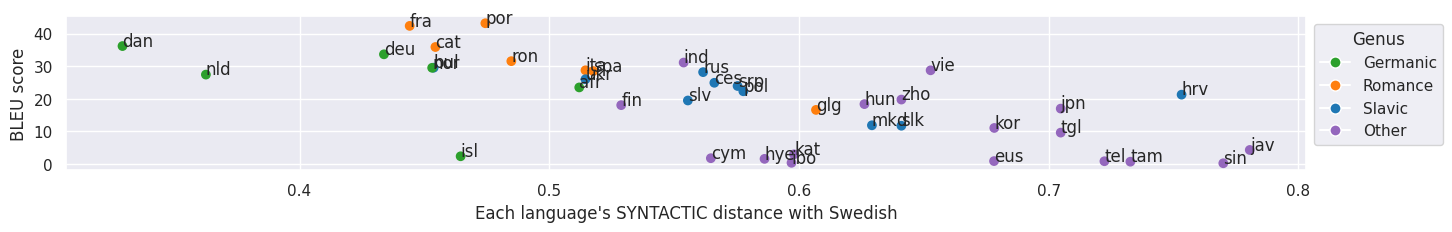}
    \caption{Scatter plot for \textbf{inllama} and \textbf{outllama} languages against the SYNTACTIC distance to \textbf{Swedish}. The correlation score is -0.67 and the p-value is \num{3.16e-06}. The negative correlation here implies that the smaller the SYNTACTIC distance of a language to Swedish, the better is its MT performance}
    \label{fig:scatter_syntactic_Swedish}
\end{figure*}
We consider several language subsets. For every language subset, we calculate the Pearson correlation score between the linguistic similarity scores of each language in the subset to a language in \textbf{inllama} and their respective translation scores. We assume that a certain language is important if we observe a positive correlation. For example, consider the language \textbf{A} and the language subset \{\textbf{B}, \textbf{C}, \textbf{D}, \textbf{E}\}. When the similarity of \textbf{A} with each language in the subset \{\textbf{B}, \textbf{C}, \textbf{D}, \textbf{E}\} and the respective machine translation scores for \{\textbf{B}, \textbf{C}, \textbf{D}, \textbf{E}\} exhibit a positive correlation, i.e. the closer they are to \textbf{A} the better their machine translation scores, \textbf{A} is deemed as a valuable language and is therefore hypothesized to be more optimal for a central language when developing multilingual language models. \textbf{A} is checked for each language in \textbf{inllama}. Similarity scores are calculated on five dimensions: GENETIC, GEOGRAPHICAL, INVENTORY, PHONOLOGY, and SYNTACTIC as per the URIEL typological database \cite{littell-etal-2017-uriel}. We exclude the FEATURAL distances to focus on each dimension as FEATURAL distances are combinations of all the other feature distances\footnote{For more detailed explanation of these distances, consult \url{https://www.cs.cmu.edu/~dmortens/projects/7_project/}}. Language subsets considered are \textbf{inllama} languages only, \textbf{outllama} languages only, both \textbf{inllama} and \textbf{outllama} languages, only \textbf{Germanic} languages, only \textbf{Romance} languages, only \textbf{Slavic} languages, and languages belonging to \textbf{Other genera}.
\begin{table}
\small
    \centering
    \begin{tabularx}{\columnwidth}{cX}
        \hline
         \textbf{Model}& \textbf{Prompt}\\
         \hline
         Non-chat& [SRC]: [src-sentence] \\
         & [TGT]: [tgt-sentence] \\
         &...\\
        &[SRC]: [src-sentence]\\
        &[TGT]:\\
         Chat& This is an English to [TGT] translation, please provide the [TGT] translation for these sentences:\\
        &[SRC]: [src-sentence] [TGT]: [tgt-sentence]\\
        &[SRC]: [src-sentence] [TGT]: [tgt-sentence]\\
        &...\\
        &Please provide the translation for the following sentence.\\
        &Do not provide any explanations or text apart from the translation.\\
        &[SRC]: [src-sentence] [TGT]: [tgt-sentence]\\
        &[TGT]\\
         \hline
    \end{tabularx}
    \caption{Prompts used in our experiments to translate languages using the non-chat and chat versions of Llama2} 
    \label{tab:prompts}
\end{table}
\begin{table}
    \small
    \centering
    \begin{tabular}{lcc} 
    \hline
         \textbf{Languages in outllama}&\textbf{BLEU}&\textbf{COMET-22}\\
    \hline
         Afrikaans&  23.52&0.74\\
         Galician&  16.62&0.76\\
         Macedonian&  11.90&0.67\\
         Slovak&  11.77&0.68\\
         Armenian&  \textbf{1.6}&0.31\\
         Basque&  \textbf{0.91}&0.33\\
         Georgian&  \textbf{2.99}&0.31\\
         Icelandic&  \textbf{2.39}&0.35\\
         Igbo&  \textbf{0.39}&0.41\\
         Javanese& \textbf{4.33}&0.59\\
         Sinhala& \textbf{0.25}&0.29\\
         Tagalog& \textbf{9.65}&0.60\\
         Tamil& \textbf{0.73}&0.30\\
         Telugu& \textbf{0.87}&0.33\\
         Welsh& \textbf{1.8}&0.35\\
    \hline
    \end{tabular}
    \caption{Llama2-7B one-shot translation results for languages in \textbf{outllama}. Languages with results in boldface are considered \textbf{unlearned} languages}
    \label{tab:outllama-res}
\end{table}
\section{Results and Analysis}
\subsection{Machine Translation Evaluation Results}
One-shot 7B Llama2 translation results are presented in Table \ref{tab:inllama-languages} and Table \ref{tab:outllama-res}. From Table \ref{tab:inllama-languages}, we observe that none of the languages included in \textbf{inllama} produce a BLEU score below 10. This suggests that we can reasonably assume that Llama2 is capable of translating into all the languages it has encountered during training. However, many languages in \textbf{outllama} yield a BLEU score under 10, this is expected as Llama2 is presumably unfamiliar with these languages. On the other hand, we hypothesize that there are two possibilities for the high-performing \textbf{outllama} languages; (1) those languages are indeed included in the training data i.e. included in the 0.005\% of the training data, or (2) similar languages in \textbf{inllama} indeed boosted their performance.

We move forward with languages in \textbf{outllama} that yield a BLEU score below 10 and experiment with other variations of Llama2. We explore the effect of scale, chat version, and adding shot count and present the results in Table \ref{tab:outllama-various}. Due to our limited compute resources we excluded the 70B and 70B-chat versions of Llama2.

\paragraph{Scaling up the model enhances translation ability. However, improvements from instruction-tuning and adding shot count remain inconclusive.} Results presented in Table \ref{tab:outllama-various} demonstrate that the 13B versions of Llama2 outperform the smaller 7B versions for all \textbf{unlearned} languages. However, larger models do not seem to yield the same number of gains for every language. In best cases, 13B models increase on average as high as 2.53 BLEU with a standard deviation of 1.64. For instruction-tuning (chat) models, we observed both performance increase and decrease. The best improvements are observed in Igbo and Javanese, which improves as much as 3.16 and 2.87 respectively, and a decrease is observed in Tagalog, which performs worse on chat models with a decrease as severe as 2.64. Adding the shot count generally improves performance although it is less drastic than model scale and instruction-tuning with a mean increase of 0.47 and 0.08 for non-chat and chat Llama-13B respectively. While these model variations appear to enhance Llama2's capacity to translate into some languages greatly, there are languages where the prospects are limited. For instance, for Sinhala and Tamil, scaling up the model/adding shot count/using chat models results in less than 1 BLEU score increase.

\begin{table*}
    \small
    \centering
    \begin{tabular}{lcccccccc}
    \hline
         \textbf{Language}&  
         \textbf{7B 1S}&
         \textbf{7B 5S}&  \textbf{7B-chat 1S}&  \textbf{7B-chat 5S}&  \textbf{13B 1S}& \textbf{13B 5S}& \textbf{13B-chat 1S}&\textbf{13B-chat 5S}\\
    \hline
         Armenian& 1.6& 1.95&  2.26&  2.43&  2.52&  \textbf{3.03}& 2.89&\textbf{3.03}\\
         Basque& 0.91& 1.08&  2.98&  3.11&  1.52&  1.9& 3.72&\textbf{3.88}\\
         Georgian&  2.99&3.44&  4.41&  4.7&  5.57&  \textbf{6.19}& 5.97&5.88\\
         Icelandic&  2.39&3.06&  3.9&  3.86&  4.72&  5.21& \textbf{5.24}&5.04\\
         Igbo& 0.39& 0.59&  1.77&  2.04&  0.56&  0.67& \textbf{3.72}&3.49\\
         Javanese&  4.33&3.71&  4.94&  5.06&  3.15&  3.76& 5.92&\textbf{6.63}\\
         Sinhala& 0.25& 0.38&  0.57&  0.52&  0.48&  \textbf{0.63}& \textbf{0.63}&0.62\\
         Tagalog&  9.65& 10.98&  10.8&  10.97&  16.1&  \textbf{16.91}& 13.82&14.27\\
         Tamil&  0.73&1.01&  0.82&  1.09&  1.79&  \textbf{2}& 1.7&1.56\\
         Telugu&0.87& 1.04& 1.02& 0.86& 2.29& \textbf{2.45}& 1.77&1.68\\
         Welsh&1.8&å 2.37& 4.38& 3.93& 5.68& \textbf{6.8}& 6.45&6.6\\
    \hline
    \end{tabular}
    \caption{BLEU scores with various Llama2 versions and shot count for languages considered \textbf{unlearned} by Llama2 (Table \ref{tab:outllama-res}). \textbf{1S}/\textbf{5S}=one-shot/five-shot. Best result for each language is bolded.}
    \label{tab:outllama-various}
\end{table*}

\subsection{Language Importance Analysis}
We use the results from Table \ref{tab:inllama-languages} and Table \ref{tab:outllama-res} for the linguistic proximity analysis. We first retrieve pre-computed distances\footnote{\url{http://www.cs.cmu.edu/~aanastas/files/distances.zip}} from the URIEL database and retrieve only the distances between the languages we are translating into and the languages reported in Llama2. Self or identity distances e.g. Igbo-to-Igbo distance are excluded in the Pearson correlation calculation. This correlation analysis aims to model the linear relationship between language proximity and machine translation scores to identify languages whose data may be beneficial for multilingual training.

\begin{figure}[H]
    \centering
    \includegraphics[width=1\linewidth,height=0.92\textheight,keepaspectratio]{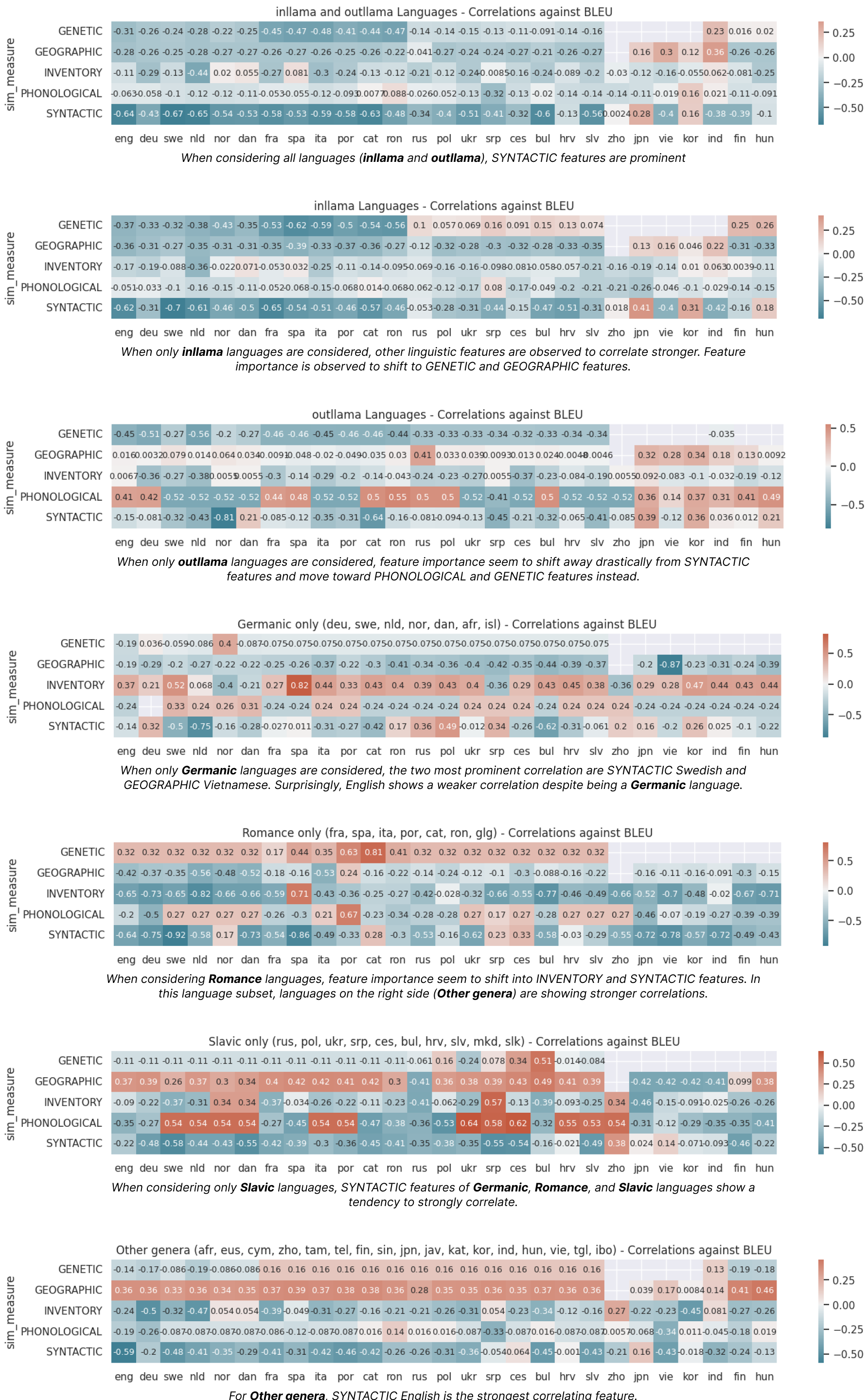}
    \caption{Heatmaps of correlations between linguistic distances with BLEU scores of the Llama2-7B one-shot prompting setup (language subset considered is written above each heatmap)}
    \label{fig:heatmap-bleu}
\end{figure}

\begin{figure}[H]
    \centering
    \includegraphics[width=1\linewidth,height=0.92\textheight,keepaspectratio]{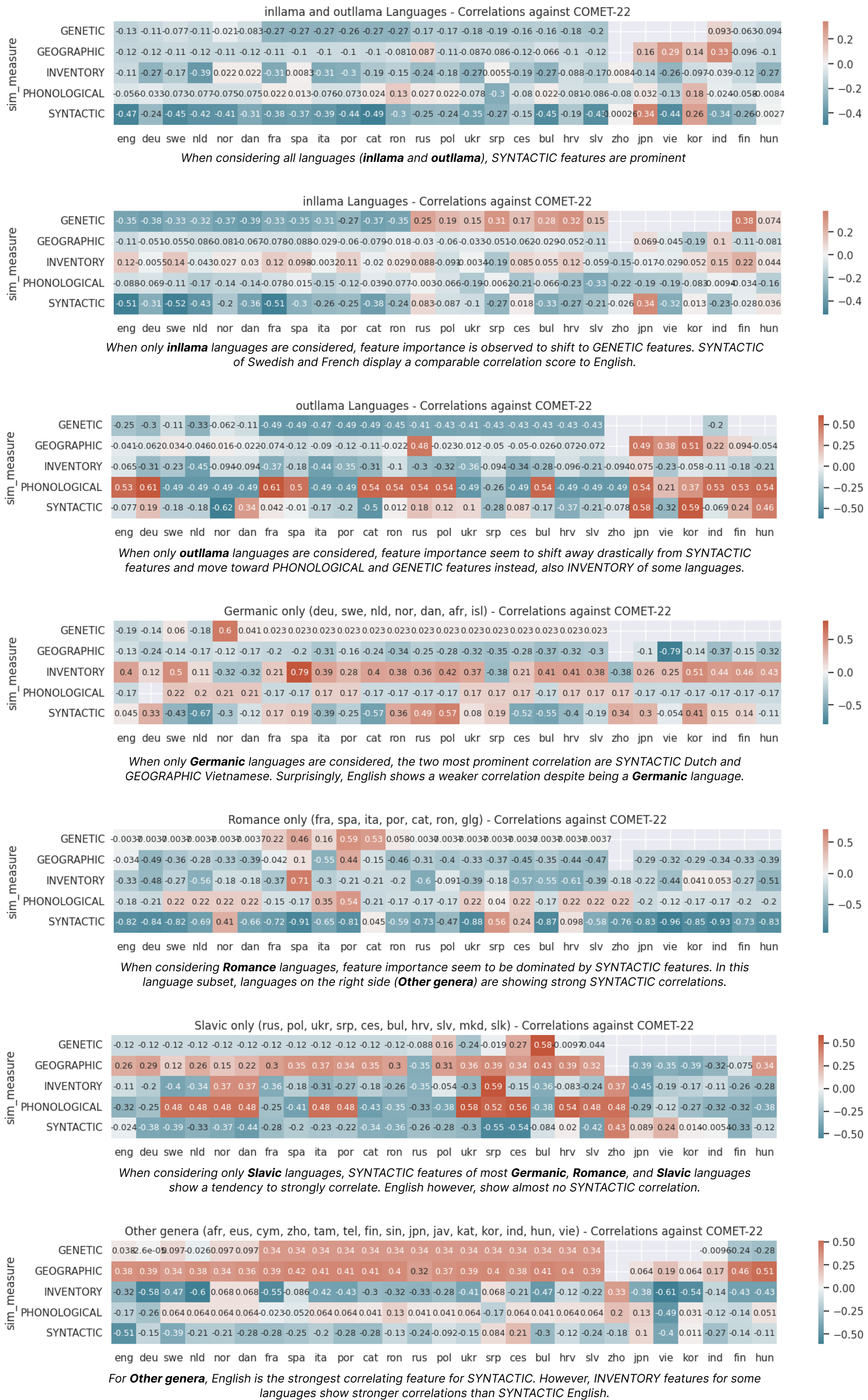}
    \caption{Heatmaps of correlations between linguistic distances with COMET-22 scores of the Llama2-7B one-shot prompting setup (language subset considered is written above each heatmap)}
    \label{fig:heatmap-comet}
\end{figure}

We present our analysis as heatmaps in Figure \ref{fig:heatmap-bleu} and \ref{fig:heatmap-comet} for correlations with BLEU and COMET-22 respectively. To help understand where each number came from in the heatmap, a scatter plot visualization for SYNTACTIC distance to Swedish for the combined \textbf{inllama} and \textbf{outllama} language subset against BLEU scores is presented in Figure \ref{fig:scatter_syntactic_Swedish} as an example. We create several different heatmaps according to the subset considered. It is important to highlight that \textit{distance} is used as a similarity score. Therefore, 
a negative correlation between linguistic distance and MT scores would imply that the closer (i.e., the \textit{smaller} the linguistic distance) a language is to this language, the higher the MT score is likely to be. 
In addition, since Wikipedia is a permanent fixture of LLMs' training data, we observe that there is a positive correlation between MT scores and Wikipedia article counts\footnote{Retrieved from \url{https://meta.wikimedia.org/wiki/List_of_Wikipedias_by_language_group} on October 2023}, as high as \textbf{0.64} using BLEU and \textbf{0.55} using COMET-22.

\paragraph{Syntactic similarity may be an important feature, but other linguistic dimensions can be too.} When including every language, i.e. the \textbf{inllama} and \textbf{outllama} subset, BLEU complemented with COMET-22 scores show consistently strong correlations with syntactic features, especially with Germanic and Romance languages. This finding may not be particularly groundbreaking, as we already understand that the languages in \textbf{inllama} predominantly belong to these language genus. However, when considering only \textbf{outllama} language subset, translation performance seems to have higher correlations (either positive or negative) with GENETIC and PHONOLOGICAL distances. 
When considering languages in \textbf{outllama}, only SYNTACTIC similarities to certain languages e.g. Norwegian and Catalan display a strong correlation with MT performances. Correlation with features other than SYNTACTIC is also observed when considering languages in \textbf{other genera}, in which the proximity with the INVENTORY feature of Vietnamese, Dutch, German, and French are shown to correlate with COMET-22 scores.

\paragraph{English is not always the most syntactically important.} 
When considering languages from \textbf{other genera} English demonstrates the most substantial syntactic correlation with MT performance, although there are other languages, such as Swedish and Vietnamese, that also display some degree of correlation. However, despite having the highest amount of training data, English is often not in the first place when considering languages by genus (e.g., \textbf{Germanic}, \textbf{Slavic}, and \textbf{Romance}). Similar to when we observe that syntactic proximity to Norwegian and Catalan have higher correlations with MT scores than syntactic proximity to English when considering only \textbf{outllama} languages, this phenomenon is accentuated when calculating correlations by genus. Among \textbf{Germanic} languages, syntactic proximity to English surprisingly shows little to no correlation with either BLEU or COMET-22 scores. Instead, \textbf{Germanic} languages' MT scores appear to correlate more with syntactic proximity to Dutch, Swedish, Catalan, and Bulgarian. This is also observed in \textbf{Slavic} languages where the MT scores generally correlate with syntactic proximity to most Germanic and Romance languages \textit{except English}. With \textbf{Slavic} languages, syntactic proximity to English has the lowest correlation on BLEU and almost no correlation on COMET-22 scores.
Finally, when focusing exclusively on \textbf{Romance} languages, it is interesting to observe that proximities to languages situated on the right side of the heatmaps i.e. \textbf{other genera}, exhibit higher correlations while they show no correlation when only considering other language subsets (Figure \ref{fig:heatmap-bleu} and \ref{fig:heatmap-comet}).

\onecolumn

\newpage
\afterpage{\clearpage}
\twocolumn

\section{Related Work}
Our work aligns with previous studies that assess LLMs for translation, resembling the work by \citet{hendy2023good} and \citet{robinson2023chatgpt}. We aim to extend such evaluations further by investigating the influence of the languages included in the training data of the model, which was previously underexplored due to the lack of transparency of LLMs used.
Our method of analysis, similar to the work of \citet{robinson2023chatgpt}, investigates feature importance. Our objective is to extend that exploration by encompassing other linguistic features obtained from the URIEL typological database \cite{littell-etal-2017-uriel}.  We are interested in the phenomenon observed in the work of \citet{lin-etal-2019-choosing} which shows that however important dataset statistics are compared to linguistic features, there are cases where using them alone to choose transfer languages results in poor performance. This phenomenon drove us to conduct a more comprehensive exploration of linguistic features.

\section{Conclusion}
We provide a comprehensive evaluation of machine translation in Llama2 for languages seen or unseen in its training data. In this work, we provide English$\rightarrow$X machine translation scores of Llama2 7B for 26 languages reported to be in the training data of Llama2 models. We also evaluated 15 additional  languages that are not reported to be in Llama2 training data using the 7B, 7B-chat, 13B, and 13B-chat Llama2 models. Our results show that Llama2 is capable of translating into languages it is unfamiliar with, although this phenomenon is observed only in some languages. We demonstrate that model scaling has the most substantial impact when compared to instruction tuning and adding shot count, whose improvements vary by language. We also modeled the linear relationship of linguistic distances and translation quality through correlation scores and revealed that syntactic similarity is not the only feature that displays strong correlations with machine translation scores. Furthermore, despite English having the most training data, there are other languages (e.g. Swedish, Catalan) whose linguistic distances exhibit comparable correlation scores to English albeit having much fewer training data. Our findings pose a unique perspective on the current landscape of language models, suggesting that the prevailing focus on English-centered models may not be the most optimal setup for multilingual models. We hope to open doors toward more effective and training-data-efficient multilingual systems that are shaped by languages other than English, thus promoting digital language equality and sustainability.

\section*{Limitations}
Our research heavily depends on the language distances obtained from the URIEL typological database, as introduced by \citet{littell-etal-2017-uriel}. The original authors noted that many languages in the database may have missing features, which means the accuracy of our findings is constrained by the methods used to compensate for these missing features. Our evaluation with the COMET-22 metric is only done for languages supported in their models. However, the model may not be equally reliable for all languages, thus the COMET-22 correlations are only as accurate as the COMET-22 model. Furthermore, there are other ways to model the relationship between language feature distances and machine translation scores. We leave such investigations for future work. We also left out positively correlated features in our analysis as they are not readily interpretable in the context of our analysis.

In an ideal scenario, it would be advantageous to include all languages from the FLORES-200 benchmark and all available versions of Llama2 and other multilingual models to provide more evidence of the effectiveness of scaling parameter count and the overall generalizability of our findings. Unfortunately, our research is constrained by limited computational resources, preventing us from achieving this comprehensive coverage. We exclude X$\rightarrow$English translation directions as Llama2 is likely trained on English Wikipedia. We also exclude prompting languages in \textbf{outllama} using various dictionary-based prompting techniques due to the challenging work required to collect accurate dictionary entries for low-resource languages. However, we leave this for future work.

We are also aware that the chat versions of Llama2 have been intentionally trained to prevent the generation of harmful or toxic content, and this protective design may affect the quality of translations. Moreover, the chat versions of the model generate numerous artifacts in addition to the translated sentences. We have made diligent efforts to automate the output parsing process to ensure that metrics are calculated fairly. The task of human evaluation and manual parsing of the outputs is left for future work.

\section*{Acknowledgements}
Authors from Indonesian institutions are supported by the Indonesian Ministry of Education, Culture, Research, and Technology (MoECRT) ACE Open Research program. This work is also supported in part by the Indonesia-US Research Collaboration in Open Digital Technology grant funded by the Indonesian Ministry of Education,
Culture, Research, and Technology. The authors would like to thank the Indonesian government for their funding and Boston University for providing essential computing resources through the Shared Computing Cluster (SCC).

\section{Bibliographical References}\label{sec:reference}

\bibliographystyle{lrec-coling2024-natbib}
\bibliography{lrec-coling2024-example}

\section{Language Resource References}
\label{lr:ref}
\bibliographystylelanguageresource{lrec-coling2024-natbib}
\bibliographylanguageresource{languageresource}
\end{document}